\documentclass{article} 
\usepackage{colm2024_conference}
\usepackage{subfigure}
\usepackage{microtype}
\usepackage{times}
\usepackage{latexsym}
\usepackage{graphicx}
\usepackage{multicol}
\usepackage{multirow}
\usepackage{adjustbox}
\usepackage{comment}
\usepackage{caption}
\usepackage{mathtools}
\usepackage{amsmath}
\usepackage{algorithm}
\usepackage{url}
\usepackage{subcaption}
\usepackage{algpseudocode}
\algblock{Input}{EndInput}
\algnotext{EndInput}
\algblock{Paramters}{EndParameter}
\algnotext{EndParameter}

\usepackage[T1]{fontenc}

\usepackage[utf8]{inputenc}

\usepackage{microtype}

\usepackage{inconsolata}

\title{Leveraging the Power of LLMs: A Fine-Tuning Approach for High-Quality Aspect-Based Summarization}

\author{$^1$Ankan Mullick \qquad $^1$Sombit Bose \qquad $^1$Rounak Saha \qquad $^2$Ayan Kumar Bhowmick \qquad \\{\bf $^2$ Aditya Vempaty \hspace{1.5mm} $^1$Pawan Goyal \hspace{1.5mm}  $^1$ Niloy Ganguly \hspace{1.5mm}  $^2$Prasenjit Dey \hspace{1.5mm}  $^2$Ravi Kokku} \\ \texttt{\{ankanm, sbcs.sombit, runk20\}@kgpian.iitkgp.ac.in}\\ \texttt{\{pawang, niloy\}@cse.iitkgp.ac.in}, \\ \texttt{\{ayan, aditya, prasenjit, ravi\}@merlyn.org}\\ $^1$Computer Science and Engineering Department, IIT Kharagpur, India.  $^2$Emergence AI}

\begin{document}


\maketitle
\begin{abstract}
The ever-increasing volume of digital information necessitates efficient methods for users to extract key insights from lengthy documents. Aspect-based summarization offers a targeted approach, generating summaries focused on specific aspects within a document. Despite advancements in aspect-based summarization research, there is a continuous quest for improved model performance. Given that large language models (LLMs) have demonstrated the potential to revolutionize diverse tasks within natural language processing, particularly in the problem of summarization, this paper explores the potential of fine-tuning LLMs for the aspect-based summarization task. We evaluate the impact of fine-tuning open-source foundation LLMs, including Llama2, Mistral, Gemma and Aya, on a publicly available domain-specific aspect based summary dataset. We hypothesize that this approach will enable these models to effectively identify and extract aspect-related information, leading to superior quality aspect-based summaries compared to the state-of-the-art. We establish a comprehensive evaluation framework to compare the performance of fine-tuned LLMs against competing aspect-based summarization methods and vanilla counterparts of the fine-tuned LLMs. Our work contributes to the field of aspect-based summarization by demonstrating the efficacy of fine-tuning LLMs for generating high-quality aspect-based summaries. Furthermore, it opens doors for further exploration of using LLMs for targeted information extraction tasks across various NLP domains.

\end{abstract}

\section{Introduction}
The ever-growing volume of information in various digital formats presents a significant challenge for users who need to efficiently extract key insights from large documents. Automatic text summarization has emerged as a valuable tool to address this challenge, providing concise and informative representations of textual documents - \cite{el2021automatic,gambhir2017recent,tas2007survey}. While traditional summarization techniques aim to capture the overall gist of a document, aspect-based summarization offers a more focused approach.

Aspect-based summarization goes beyond generic summarization by targeting specific aspects or topics within a document - \cite{frermann2019inducing, coavoux2019unsupervised, mukherjee2020read}. This targeted approach is particularly valuable for large documents, such as research papers, product reviews, or news articles, where specific information about certain aspects might be crucial for the reader.
Aspect-based summarization allows users to delve deeper into a document by quickly providing summaries that cater to their specific information needs. For instance, consider a researcher reviewing a medical study. Using aspect-based summarization, they can prioritize summaries that highlight the methodology and results sections. Conversely, a customer reading online reviews for a new phone might prioritize summaries emphasizing aspects like battery life or camera performance - \cite{li2020aspect, kunneman2018aspect}. 

Hence, effective generation of aspect-based summaries presents a unique challenge. Unlike generic summarization, which focuses on capturing the overall gist of a document, aspect-based summarization requires models to not only comprehend the document's content but also identify and extract information pertinent to specific aspects. This necessitates models that can not only understand the semantics of the text but also possess the ability to discern and prioritize aspect-related information.

Despite the significant strides made in aspect-based summarization research, there remains an ongoing quest for models capable of generating even higher quality summaries. While several state-of-the-art methods like Falcon, BART, Pegasus, T5, and LED [\cite{lewis2020bart, penedo2023refinedweb, wan2022factpegasus, guo2022longt5}] among others have yielded promising results, there is a continual exploration for novel approaches that can elevate the quality of aspect-based summaries. This quest motivates the exploration of new approaches, such as fine-tuning large language models (LLMs) (\cite{huang2022large, ding2023parameter}) for the task of aspect-based summarization.

LLMs represent a transformative paradigm shift in natural language processing (NLP). These powerful models, trained on massive datasets of text and code, have demonstrated exceptional capabilities in various NLP tasks, including text generation, translation, and question answering among others (\cite{wei2022emergent, hoffmann2022training}). The ability of LLMs to capture intricate linguistic patterns and relationships within text(\cite{yang2023exploring}) makes them a compelling candidate for enhancing the performance of aspect-based summarization task.

In this paper, we aim to study the impact of finetuning LLMs (\cite{yang2024unveiling}) for the task of aspect-based summarization and demonstrate the improvement in the quality of generated aspect-based summaries over vanilla LLMs. Our work centers around the concept of fine-tuning recent open-source foundation LLMs, including Llama2 (\cite{touvron2023llama}), Mistral (\cite{jiang2023mistral}), Gemma (\cite{team2024gemma}) and Aya (\cite{ustun2024aya}).
Precisely, we investigate the potential of fine-tuning such open-source foundation LLMs on a dataset specifically tailored for the task of aspect-based summarization. By fine-tuning these LLMs on aspect-based summarization datasets, we aim to equip them with the necessary expertise to effectively identify, extract, and generate summaries that focus on user-specified aspects within a document such that the fine-tuned LLMs can achieve superior performance compared to existing methods. 
In this paper, we seek to address the following research questions through making contributions related to the field of aspect-based summarization:

\noindent 1. Does fine-tuning LLMs provide a significant benefit for aspect-based summarization tasks?

\noindent 2. How effective are fine-tuned LLMs compared to vanilla LLMs and other state-of-the-art methods for aspect-based summarization?

\noindent 3. Does the effectiveness of fine-tuning LLMs vary depending on the base model architecture?

\noindent 4. How robust is the fine-tuned LLM for variations in dataset and domains for aspect-based summarization?

\section{Related Work}
In this section, we perform a survey of the state-of-the-art literature on summarization and discuss the literature on different types of summarization as follows:

\subsection{Generic Summarization} 

We focus on brief literature survey on generic summarization, which encompasses a broad approach to summarizing text without focusing on specific aspects, queries, or goals, using abstractive or extractive approaches. 
Among abstractive approaches, \cite{chopra2016abstractive} introduced an abstractive summarization model using attentive recurrent neural networks and discuss the challenges of generating coherent and informative summaries while avoiding redundancy. Based on pointer-generator network framework, \cite{see2017get} presents a model that combines extractive and abstractive techniques for summarization by effectively incorporating source information into the generated summaries. On the other hand, among purely extractive approaches, earlier researchers used graph-based approaches like TextRank (\cite{mihalcea2004textrank}) and LexRank (\cite{erkan2004lexrank}).   

\subsection{Aspect-based Summarization}
\cite{hayashi2021wikiasp} employed a method for aspect-based summarization focusing on multiple domains while (\cite{coavoux2019unsupervised}) focused on aspect-based multi-document abstractive summarization with an unsupervised approach. Few works have also explored domain-specific aspect-based summarization such as (\cite{mukherjee2020read}) that focus on data from tourist review domain and (\cite{akhtar2017aspect}) that focus on dataset of hotel reviews. (\cite{tang2016aspect}) developed a deep memory network for aspect-level sentiment classification, emphasizing the extraction of aspects within a document and these are relevant for aspect-based summarization. Again, (\cite{wang2016attention}) proposed an attention-based LSTM model which helps identify and emphasize important aspects and these are used for aspect-based summarization.
\subsection{Use of LLMs for summary evaluation} 
LLMs are recently emerging as alternatives to traditional metrics and human evaluation for evaluating NLP tasks. 
Recent work has explored LLM-based NLG evaluation methods (\cite{gao2024llm}) while (\cite{chan2023chateval}) assessed the quality of generated responses from different models on open-ended
questions. (\cite{zhou2023don}) have proposed guidelines for LLM use for evaluations while few works have proposed techniques to improve LLM evaluation performance [\cite{hasanbeig2023allure,liu2023calibrating}]
and (\cite{huang2023can}) have also investigated the explainability of LLMs in evaluation contexts.

In this paper, our focus is on analysing the impact of fine-tuning open-source foundation LLMs on the performance of the aspect-based summarization task and determine the type of LLMs that can help to generate high quality aspect-based summaries either using the pre-trained version or after fine-tuning on relevant datasets. We also use LLMs (GPT4) to evaluate summaries on different conventions.


\section{Dataset}
We leverage the publicly available benchmark dataset, Open Aspect-based Summarization (OASUM)~\cite{yang2022oasum}, for both fine-tuning open-source foundation LLMs and evaluating their performance. OASUM offers a rich collection of over 3.6 million document-aspect-summary triplets, featuring diverse aspects across various domains\footnote{\url{https://github.com/tencent-ailab/OASum}}. There are 1M unique aspects in the entire dataset. The average token count for the documents and aspect-based summaries are $1612$ and $40$ respectively.

\begin{table*}[!ht]
    \centering
    \begin{adjustbox}{width=0.95\columnwidth}
    \begin{tabular}{|c|c|}
        \hline
        Domain & Aspect set \\
        \hline
        HealthCare & Death, Diagnosis, Differential diagnosis, Diagnosis-Classification \\
        \hline
        Education & History, Geography, Taxonomy, Education\\
        \hline
        Life and Career & Career, Political Career, Personal Life, Life and career\\
        \hline
         Music & Production, Composition, Soundtrack, Track Listing\\
        \hline
    \end{tabular}
    \end{adjustbox}
    \caption{Domain-wise breakdown of aspects in OASUM dataset}
    \label{tab:domain_aspect_set}
\end{table*}

\textbf{Data Preprocessing and Variations:} To facilitate targeted training and analysis, we prepared several variations of the OASUM dataset:

\noindent \textbf{1. Domain-Wise Split:} We selected 16 aspects from four popular domains (Healthcare, Music, Education, Life \& Career) resulting in a domain-specific dataset of 14,279 training instances.

\noindent \textbf{2. High-Frequency Aspects:} We created the variation \textit{OASUM-Hi} by choosing the top-50 most frequent aspects (based on document count) and randomly selecting 1,000 documents for each. This dataset investigates the impact of fine-tuning on well-represented aspects.
    
\noindent \textbf{3. Low-Frequency Aspects:} In contrast, the variation \textit{OASUM-Lo} focuses on the long tail of the dataset. We selected the 50 least frequent aspects ($1-4$ document occurrences) with 1,000 documents each. This explores fine-tuning performance on less common aspects (aka long-tails).
    
\noindent \textbf{4. Random Aspect Selection:} The variation \textit{OASUM-Ra} comprises a randomly selected set of 50,000 document-aspect-summary triplets for a domain-agnostic evaluation.

Table~\ref{tab:dataset} summarizes the key statistics for each dataset variation, including aspects, training/validation/test split sizes.
\begin{table}[!th]
    \centering
    \begin{adjustbox}{width=0.65\columnwidth}
    \begin{tabular}{|c|c|c|c|c|}
        \hline
         Dataset & Aspect & Train & Validation & Test \\\hline
        OASUM-domain wise & 16 & 14279 & 500 & 2544\\\hline
        OASUM-Hi & 50 & 50000 & 500 & 500 \\\hline 
        OASUM-Lo & 8995 & 50000 & 500 & 500\\\hline
        OASUM-Ra & 7320 & 50000 & 500 & 500 \\\hline
    \end{tabular}
    \end{adjustbox}
    \caption{Different OASUM Datasets distribution}
    \vspace{-3mm}
    \label{tab:dataset}
\end{table}

\section{Proposed Framework}
In this section, we detail our framework for fine-tuning open-source foundation LLMs on the \textit{OASUM} dataset to obtain corresponding fine-tuned domain-specific LLMs specialized for the downstream task of aspect-based summarization. We describe the fine-tuning process, the LLM architectures employed, and the baseline models used for comparison.

\subsection{Model architecture for fine-tuning LLMs} Our training process consists of employing different open-source foundation LLMs for fine-tuning on the training set of \textit{OASUM} dataset described above. Specifically, we leverage supervised fine-tuning (\cite{zhang2023balancing}) on the \textit{OASUM} training dataset to transform pre-trained foundation LLMs into domain-specific models suited to perform aspect-based summarization. This involves utilizing prompt-completion pairs to guide the pre-trained models towards generating aspect-based summaries. Each training instance comprises a document paired with an instruction to generate a summary based on a specific aspect. The corresponding completion is the relevant aspect-based summary.

To enhance the fine-tuning process, we incorporate advanced techniques like Quantized Low-Rank Adaptation (QLoRA) \cite{dettmers2023qlora} and PEFT (Parameter-Efficient Fine-Tuning)~\cite{fu2023effectiveness} to optimize training efficiency. Following fine-tuning, these models (referred to as "*FT") acquire the ability to generate aspect-based summaries for corresponding documents based on the specified aspect within the prompt.
Following is a summary of the open-source foundation LLMs we fine-tuned on OASUM:

\noindent \textbf{1. Llama2:}\footnote{\url{https://ai.meta.com/llama/}} We use two different versions of Llama2 - vanilla: with sizes of 7b, 13b and 70b~(\cite{touvron2023llama}) and fine-tuned: using models Llama2-7b and 13b. We have referred the Llama2-7b and Llama2-13b fine-tuned version as Lm7b-FT and Lm13b-FT.
    
\noindent \textbf{2. Mistral:} We fine-tuned the Mistral-7b decoder-only Transformer model (\cite{jiang2023mistral}) from Mistral AI, obtaining Mistral-7b-FT (abbreviated as Mis7b-Va for vanilla and Mis7b-FT for finetune).
    
\noindent \textbf{3. Gemma:} We use Gemma which is a family of lightweight, state-of-the-art open models (\cite{team2024gemma}) developed by Google DeepMind from the same technology used to create the Gemini models. Specifically, we finetune the Gemma-2b version to obtain the finetuned version referred to as Gemma-FT.
    
\noindent \textbf{4. Aya:} We use the Aya Model (\cite{ustun2024aya}), a massively multilingual 13 billion parameter language model capable of following instructions in 101 languages that is developed by Cohere and fine-tune the pre-trained version to obtain the fine-tuned model referred to as Aya-FT. 

For performance comparison, we also include the vanilla pre-trained versions of each LLM (referred to as "*VA"). These include Llama2-7b-VA (Lm7b-VA), Llama2-13b-VA (Lm13b-VA), Llama2-70b-VA (Lm70b-VA), Mistral-7b-VA, Gemma-VA, and Aya-VA. 


\subsection{Baseline models} We use the following state-of-the-art competing baselines for comparing the performance of aspect-based summarization task against the fine-tuned LLMs and their vanilla counterparts:

\noindent \textbf{1. LongFormer:} The Longformer (\cite{beltagy2020longformer}) is a transformer-based model designed to handle long documents efficiently using an attention pattern that effectively combines local and global information, enabling to handle long inputs. We use Longformer-base (LED-ba) and large (LED-La) model with 149 million and 439 million parameters. 
    
\noindent \textbf{2. T5 (Text-to-Text Transfer Transformer):} This model~\cite{raffel2020exploring} leverages transfer learning for summarization tasks by converting them into a text-to-text format. We fine-tune the T5-3b version (T5-FT) with 3 billion parameters to generate aspect-based summaries. 
    
\noindent \textbf{3. Flan T5:} Flan-T5~\cite{chung2022scaling} instruction fine-tuned approach highlights the benefits of fine-tuning across various models, prompting setups, and evaluation tasks. We finetune the Flan T5 XL model (Fl-T5-FT).

\noindent \textbf{4. BART (Bidirectional and Autoregressive Transformer):} This denoising autoencoder~\cite{lewis2019bart} is used for pre-training sequence-to-sequence models. We employ the instruction-prompted BART-large model with 406 million parameters, pre-trained on English and fine-tuned for summarization on the CNN Daily Mail news dataset.
    
\noindent \textbf{5. Pegasus:} We utilize the instruction-tuned Pegasus model~\cite{zhang2020pegasus} with 571 million parameters for generating aspect-based summaries.
    
\noindent \textbf{6. Falcon:} The Falcon 7b-instruction-tuned model~\cite{penedo2023refinedweb} is used for generating aspect-based summaries.
    
\noindent \textbf{7. TLDR:} We apply state-of-the-art approach `TLDR-CATTS-XSUM' (TLDR)~\cite{cachola2020tldr} for extreme summarization to obtain crisp summary of the document.   

\section{Experimental evaluation and Results}
In this section, we evaluate the performance of our different fine-tuned LLM models in terms of the quality of the generated aspect-based summaries for documents in the \textit{OASUM domain wise} test set and compare against their vanilla counterparts as well as the competing baseline models.
\subsection{Evaluation metrics and experimental settings} 
Our evaluation relies on two different approaches:

\noindent \textbf{1. Traditional:} Here we check the comptenece of different models with traditional evaluation metrics like (i) Rouge 1 (R1), Rouge 2 (R2) and Rouge L (RL) (\cite{lin2004rouge}), (ii) Meteor (Mt) (\cite{banerjee2005meteor}), (iii) Bleu (Bl) \cite{papineni2002bleu}), and (iv) BERTScore F1 (BeF1) (\cite{zhang2019bertscore}) to assess the quality of generated summaries.
    
\noindent \textbf{2. GPT-4 Critique:} Here, we use the GPT-4 LLM as a critique~\cite{valmeekam2023can, sun2024critique} to evaluate the quality of the model generated aspect-based summaries against the gold standard aspect-based summaries in the test set of the \textit{OASUM} dataset variations from different dimensions. Specifically, we provide suitable critique based prompts to GPT-4 where we evaluate the summaries based on a set of five predefined criterias (termed as \textit{GPT-4 criteria}) defined below:
    
    \textit{a. Relevance (Re):} The extent to which the generated summary is relevant to the specific aspect-based summary of the document. 
    
    \textit{b. Coverage (Cv):} The extent to which the generated aspect-based summary correctly covers all the important key points described in the gold standard aspect-based summary of the document. 
    
    \textit{c. Impurity (Im):} The extent to which the aspect-based summary does not contain information specific to any other aspect. 
    
    \textit{d. Rating (Ra):} Scores how well the summary captures the target aspect with the score reflecting if the summary is good, average or bad. A good summary is clear, concise, accurate, and engaging. An average summary conveys the main points but might lack detail. A bad summary is inaccurate, unclear, non-coherent or overly verbose. (Details are in Appendix)
    
    \textit{e. Goodness (Gd):} Extending from 4, we manually verify the goodness of the summary. 

This combined evaluation strategy allows us to assess performance from both a similarity and quality perspective, leveraging established metrics and leveraging the capabilities of GPT-4 for in-depth analysis.

\textbf{Experimental Settings:} We use 80GB A100 GPU, 210MHz clock cycle and 6 epochs for all experiments  (Details are in Appendix). We have used NLTK, Spacy, openai(version=0.28), huggingface\_hub, torch and transformers python packages for all experiments\footnote{Code/Data are in \url{http://tiny.cc/zjelxz}}.

\begin{table*}[!ht]
    \centering
    \begin{adjustbox}{width=0.95\columnwidth}
    \begin{tabular}{|c|c|c|c|c|c|c|c|c|c|c|c|c|}
        \hline
        \textbf{Model} & \textbf{R1} & \textbf{R2} & \textbf{RL}& \textbf{Mt} & \textbf{Bl}  & \textbf{BeF1} & \textbf{Re} & \textbf{Cv} & \textbf{Im} & \textbf{Ra} & \textbf{Gd}\\
        \hline
        Llama2-7b-FT & 39.4 & 23.9 & 35.9  & 32.7 & 14.7 & 80.0  & 65.8 & 45.2 & 96.6 & 55.2 & 37.7\\\hline
        Llama2-13b-FT & \textbf{41.5} & \textbf{25.9} &  \textbf{37.8} & \textbf{35.5} & \textbf{16.8} &  \textbf{80.7}  & \textbf{68.3} & \textbf{48.9} & \textbf{96.7} & 58.8 & 42.3\\\hline
        Mistral-7b-FT & 36.1 & 19.8 & 31.6 & 30.8 & 11.8 &  78.8  & 67.7 & 46.2 & 83.5 & \textbf{61.4} & \textbf{56.0}\\\hline 
        Gemma-FT & 17.3 & 2.4 & 10.9 & 8.7 & 0.7 &  62.4  & 59.7 & 37.1 & 79.0 & 48.1 & 20.0\\\hline
        Aya-FT & 22.9 & 10.6 & 20.1 & 15.9 & 4.2 &  68.2  & 35.2 & 27.0 & 57.8 & 41.1 & 40.0 \\\hline
        Falcon & 17.2 & 4.8 & 12.5 & 22.2 & 1.2 & 71.6  & 61.5 & 42.1 & 87.5 & 55.1 & 40.2\\\hline
        BART & 23.9 & 8.5  & 17.6 & 27.5 & 3.3 &  74.6  & 62.4 & 43.1 & 86.8 & 52.1 & 22.1\\\hline
        Pegasus & 19.8 & 5.5 & 14.2 & 21.9 & 1.9 &  71.9  & 50.9 & 37.0 & 87.0 & 45.5 & 30.7\\\hline
        T5-FT & 35.2 & 18.2 & 31.1 & 29.5 & 10.1 &  78.7  & 63.3 & 42.7 & 95.1 & 53.7 & 24.0\\\hline
        Fl-T5-FT & 35.8 & 19.1  & 31.6 & 30.6 & 10.9 & 79.1  & 64.4 & 44.1 & 94.8 & 54.9 & 25.5\\\hline
       LED-ba & 28.2 & 16.6 & 26.1 & 24.6 & 9.9 &  71.9  & 54.2 & 38.0 & 83.5 & 48.6 & 22.8\\\hline
       LED-la & 34.2 & 18.5 & 30.9 & 27.7 & 10.9 &  75.9  & 62.1 & 40.9 & 85.7 & 42.9 & 39.6\\\hline
       TLDR & 28.2 & 12.1 & 23.8 & 21.8 & 4.1 &  76.2  & 52.4 & 48.1 & 80.8 & 49.1 & 22.1\\\hline
    \end{tabular}
    \end{adjustbox}
    \caption{Traditional and GPT-4 based evaluation on OASUM domain-wise dataset variation}
    \vspace{-3mm}
    \label{tab:oasum_rouge}
\end{table*}

\subsection{Results and discussion}
In this section, we analyze the results presented in Table~\ref{tab:oasum_rouge} as well as Figure~\ref{fig:class_dist} based on values of traditional metrics and GPT-4 criteria respectively to understand how different models perform and gain insights into the effectiveness of fine-tuning LLMs for aspect-based summarization.

\subsubsection{How effective is fine-tuning LLMs for aspect-based summarization based on traditional evaluation metrics?}
\begin{figure}[!htb]
\centering
        \begin{subfigure}
                \centering
                \includegraphics[width=0.47\textwidth]{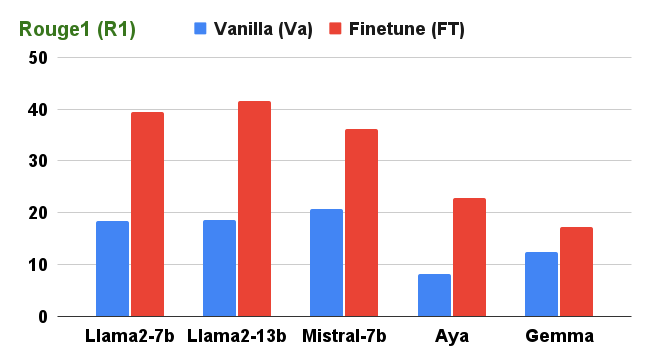}
                \label{fig:r1}
        \end{subfigure}%
        \begin{subfigure}
                \centering \includegraphics[width=.47\textwidth]{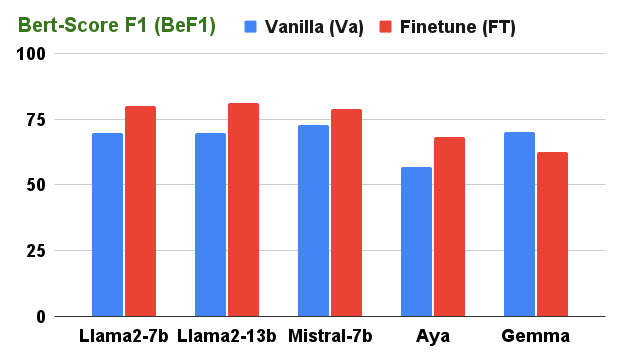}
                \label{fig:bef1}
        \end{subfigure}%
        \begin{subfigure}
                \centering
                \includegraphics[width=.47\textwidth]{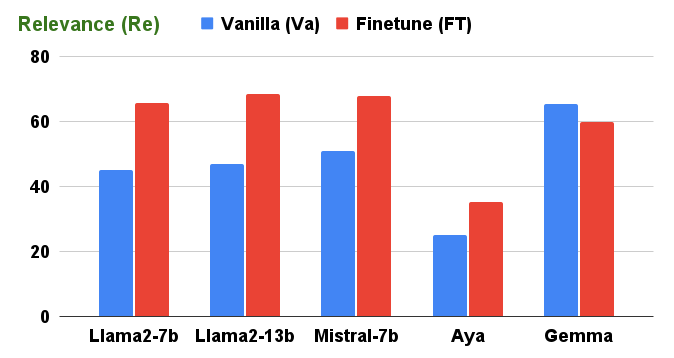}
                \label{fig:re}
        \end{subfigure}%
                \begin{subfigure}
                \centering
                \includegraphics[width=.47\textwidth]{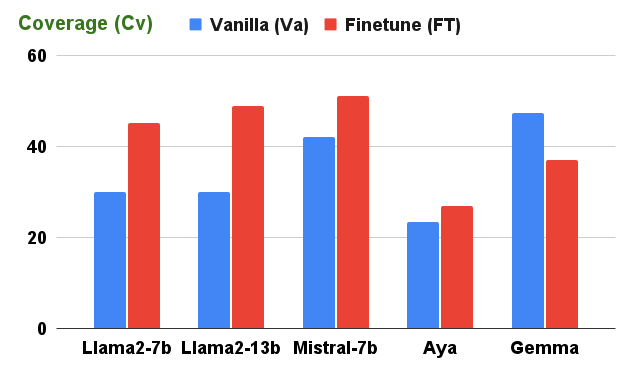}
                \label{fig:cv}
        \end{subfigure}%
        \vspace{-2mm}
        \caption{Comparison between vanilla and fine-tuned versions of different LLMs for Rouge1, Bert-Score F1, Relevance and Coverage} 
        \vspace{-6mm}
        \label{fig:class_dist}
\end{figure}

In Figure~\ref{fig:class_dist}, we can see comparison between vanilla and fine-tuned LLMs based on values for traditional metrics like ROUGE and BERTScore. Here, we can observe a significant performance boost for fine-tuned LLMs (particularly Llama2-7b-FT, Llama2-13b-FT, Mistral-7b-FT) compared to their vanilla counterparts (Llama2-7b-VA, Llama2-13b-VA, Mistral-7b-VA) across all metrics. This indicates that fine-tuning successfully tailors these models to the task of aspect-based summarization, enabling them to generate summaries that better match the gold-standard summaries in terms of n-gram overlap and semantic similarity.

Among the fine-tuned LLMs, Llama2-13b-FT consistently achieves the highest scores across all traditional metrics compared to competing baseline models (as seen from Table~\ref{tab:oasum_rouge}), suggesting that its larger parameter size provides an advantage in capturing the nuances of aspect-based information. Interestingly, among the latest released LLMs, Aya-VA demonstrates an expected performance gain upon fine-tuning, suggesting its potential suitability for aspect-based summarization tasks. However, Gemma-VA degrades in BeF1 score, highlighting the importance of model architecture and suitability for aspect-based summarization task beyond parameter size. In summary, all models might NOT gain performance upon finetuning.

\subsubsection{How does fine-tuning LLMs impact the quality of summaries based on GPT-4 critiquing?}
Table~\ref{tab:oasum_rouge} and Fig.~\ref{fig:class_dist} also unveil a deeper perspective on summary quality through the lens of GPT-4 critiquing. Here, we evaluate summaries based on five criteria: relevance, key point coverage, aspect-specificity, overall quality, and manually verified goodness.
Consistent with the traditional metrics, fine-tuned LLMs (Llama2-7b-FT, Llama2-13b-FT, Mistral-7b-FT) significantly outperform vanilla models (Llama2-7b-VA, Llama2-13b-VA) across all criteria as further supported by corresponding plots comparing vanilla and fine-tuned LLMs based on values of relevance and coverage in Figure~\ref{fig:class_dist}. This reinforces the effectiveness of fine-tuning in generating summaries that are not only similar to the gold standard but also capture the essence of the specific aspect and deliver clear, concise information.

Llama2-13b-FT achieves best performance again compared to baseline methods, achieving the highest scores in most criteria, particularly in key point coverage and overall quality (see Table~\ref{tab:oasum_rouge}). This suggests that its larger size allows for a more comprehensive understanding of the document and the target aspect, leading to summaries that effectively capture the crucial aspect-based details. However, size is not necessarily an indicator of getting the best performance since Aya is also 13b moel but has the least performance for this task among the models considered. This indicates that specific models are optimized for specific tasks. Also, similar to our observations in the previous sections, Gemma-FT has degraded performance upon fine-tuning, indicating that fine-tuning LLMs is not always a good thing for all LLMs and tasks.

\subsubsection{Which LLMs achieve the best performance on fine-tuning?}
By combining the insights from both traditional metrics and GPT-4 critiquing results, Llama2-13b-FT emerges as the clear winner for generating aspect-based summaries, consistently demonstrating superior performance in terms of similarity, key point coverage, relevance, and overall quality. Its larger parameter size appears to be instrumental in achieving this level of performance for the aspect-based summarization task, along with its superior architecture.

These findings significantly strengthen the case for fine-tuning LLMs for aspect-based summarization, for most of the base models. Fine-tuning not only improves the similarity of generated summaries to the gold standard but also enhances their ability to capture the essence of the target aspect and deliver clear, concise information. While parameter size plays a role, model architecture also plays a crucial part, as evidenced by Gemma-VA's limitations with fine-tuning not improving its performance and the marginal improvement of Aya-FT over its vanilla counterpart.
\begin{table*}[]
\begin{adjustbox}{width=0.95\columnwidth}
\begin{tabular}{|c|c|c|c|c|c|c|c|c|c|c|c|c|c|}
\hline
\textbf{Data}      &           \multicolumn{2}{c|}{\textbf{Approach}}         & \textbf{R1} & \textbf{R2} & \textbf{RL} & \textbf{Mt} & \textbf{Bl} & \textbf{BeF1} & \textbf{Re} & \textbf{Cv} & \textbf{Im }& \textbf{Ra} & \textbf{Gd} \\ \hline
\multirow{6}{*}{Hi} & \multirow{3}{*}{VA} & Lm7b-VA & 18.5 & 5.4 & 13.9 & 20.8 & 1.5 & 70.0 & 42.1& 36.5 & 55.1 & 40.1& 26.4\\ \cline{3-14} 
                  &                   & Lm13b-VA & 19.2 & 5.5 & 14.1 & 21.6 & 1.8 & 68.5 & 43.2& 37.2 & 56.3& 43.6 & 28.8\\ \cline{3-14} 
                  &                   & Mis7b-VA & 22.1 & 6.6 & 15.8 & 25.2 & 2.1 & 73.1 & 59.7 & 38.1 & 85.3 & 50.6 & 24.0\\ \cline{2-14} 
                  & \multirow{3}{*}{FT} & Lm7b-FT & 33.8 & 18.3 & 30.7 & 28.3 & 10.0 & 78.4 & 53.7 & 43.8 & 69.1 & 49.2 & 35.3 \\ \cline{3-14} 
                  &                   & Lm13b-FT & \textbf{36.9} & \textbf{21.9} & \textbf{33.0} & \textbf{30.3} & \textbf{11.7} & \textbf{81.1} & \textbf{63.2}& \textbf{44.8}& 88.1& \textbf{52.3}& \textbf{42.5}\\ \cline{3-14} 
                  &                   & Mis7b-FT & 32.4 & 15.9 & 27.6 & 26.6 & 7.1 & 78.1 & 59.4 & 42.3 & \textbf{90.3} & 47.3 & 42.0 \\ \hline
\multirow{6}{*}{Lo} & \multirow{3}{*}{VA} & Lm7b-VA & 14.3& 4.4& 10.9& 17.0& 1.0& 65.3 & 29.1& 27.5& 48.3& 30.5& 20.6\\ \cline{3-14} 
                  &                   & Lm13b-VA & 20.1 & 5.3 & 14.6 & 20.3 & 1.6 & 70.2 & 31.5 & 25.2& 48.9& 29.6& 20.1\\ \cline{3-14} 
                  &                   & Mis7b-VA & 21.5 & 5.7 & 15.2 & 20.1 & 1.6 & 73.2 & 46.0 & 30.5 & 55.6 & 42.5 & 18.0 \\ \cline{2-14} 
                  & \multirow{3}{*}{FT} & Lm7b-FT & 21.9 & 7.2 & 16.4 & 22.1 & 3.5 & 72.1 & 34.1& 32.7& 50.2& 31.7& 25.2\\ \cline{3-14} 
                  &                   & Lm13b-FT & \textbf{29.2} & \textbf{13.3} & \textbf{25.1} & \textbf{22.5} & \textbf{6.3} & \textbf{78.8} & \textbf{48.2} & \textbf{36.6}& \textbf{66.8}& \textbf{49.9}& \textbf{44.8}\\ \cline{3-14} 
                  &                   & Mis7b-FT & 25.3 & 8.8 & 20.3 & 19.4 & 2.7 & 76.1 & 47.5 & 32.3 & 62.8 & 43.8 & 42.0\\ \hline
\multirow{6}{*}{Ra} & \multirow{3}{*}{VA} & Lm7b-VA & 15.5& 4.9& 11.7& 19.7 & 1.4& 68.2 & 34.6 & 30.2& 49.0 & 32.7& 21.9 \\ \cline{3-14} 
                  &                   & Lm13b-VA & 19.6 & 5.3 & 14.2 & 21.3 & 1.6 & 70.3 & 35.2& 31.8& 52.4& 33.3& 23.3\\ \cline{3-14} 
                  &                   & Mis7b-VA & 21.8 & 5.9 & 15.6 & 24.8 & 1.8 & 70.2 & 52.2 & 30.7 & 80.9 & 39.2 & 22.0\\ \cline{2-14} 
                  & \multirow{3}{*}{FT} & Lm7b-FT & 27.8 & 13.9 & 27.2 & 26.1 & 7.8 & 73.0 & 48.9 & 35.4& 62.3& 34.3& 29.0 \\ \cline{3-14} 
                  &                   & Lm13b-FT & \textbf{30.4} & \textbf{14.6} & \textbf{28.1} & \textbf{28.3} & \textbf{9.2} & \textbf{75.3} & \textbf{55.8}& \textbf{39.0}& \textbf{88.9}& \textbf{42.5}& \textbf{33.9}\\ \cline{3-14} 
                  &                   & Mis7b-FT & 28.6 & 13.1 & 24.8 & 24.1 & 5.3 & 72.3 & 53.7 & 33.6 & 86.8 & 40.0 & 38.0\\ \hline
\end{tabular}
\end{adjustbox}
\caption{Traditional and LLM based evaluations on three different variations of OASUM dataset}
    \label{tab:oasumHi-Lo-Ran_Trad-gpt}
    \vspace{-3mm}
\end{table*}

\subsubsection{How robust is the fine-tuned LLM for variations in dataset and domains for aspect-based summarization?}
To answer this question, we pick our best perfoming fine-tuned model from the results in the previous section, the Llama2-13b-FT, and evaluate it on variations in the dataset.

\paragraph{Different Types of OASUM Data:}
To check the effectiveness of oir fine-tuned models, we experiment on different types of OASUM data: \textit{OASUM-Hi}, \textit{OASUM-Lo} and \textit{OASUM-Ra} as shown in Table \ref{tab:oasumHi-Lo-Ran_Trad-gpt}. By employing multiple dataset variations, we aim to achieve a comprehensive evaluation of fine-tuned LLMs for aspect-based summarization, taking into account various data characteristics and potential shortcomings in existing summaries. As expected, evaluation outcomes are best for \textit{OASUM-Hi} and least for \textit{OASUM-Lo} since number of aspects for \textit{OASUM-Hi} is much lesser than \textit{OASUM-Lo}. \textit{OASUM-Ra} exhibit results better than \textit{OASUM-Lo} due to presence of lesser aspects than \textit{OASUM-Lo}. Llama2-13b-FT performs best for almost all scenrios across different evaluation metrics.

\begin{table*}[!ht]
    \centering
    \begin{adjustbox}{width=0.95\linewidth}
    \begin{tabular}{|c|c|c|c|c|c|c|c|c|c|c|c|}
        \hline
        \textbf{Domain} & \textbf{R1} & \textbf{R2} & \textbf{RL}& \textbf{Mt} & \textbf{Bl} & \textbf{BeF1} & \textbf{Re} & \textbf{Cv} & \textbf{Im} & \textbf{Ra} & \textbf{Gd}\\
        \hline
        Healthcare &	32.8 &	17.3 &	29.3 &	27.4 &	9.4 & 77.8 & 69.9 & 47.0 & 97.4 & 57.0 & 42.5\\\hline
Education	& 44.9 &	28.2 &	41.1  &	38.1 &	18.2 & 81.3 & 68.2 & 51.0 & 97.6 & 58.6 & 45.3\\\hline
Life and Career &	39.4 &	23.9 &	35.5 &	32.7 &	14.1 & 80.4 & 69.9 & 48.5 & 96.7 & 58.8 & 41.1\\\hline
Music	& 41.9 & 27.6 &	38.6  &	37.7 & 20.4 & 81.0 & 66.2 & 0.47 & 94.9 & 56.6 & 40.3\\\hline
Average	& 41.5 & 25.9 &	37.8 & 35.5 & 16.8 & 80.7 & 68.3 & 48.9 & 96.7 & 58.8 & 42.3\\\hline
    \end{tabular}
    \end{adjustbox}
    \caption{Evaluations of fine-tuned Llama2-13b on different domains of OASUM}
    \vspace{-4mm}
    \label{tab:oasum-domain_rouge_gpt}
\end{table*}


\paragraph{Evaluations for different Domains:} In Table \ref{tab:oasum-domain_rouge_gpt}, we show five different traditional metric and GPT4 critique scores for the best performing Llama2-13b Finetuned model  of OASUM data for 4 different domains - Healthcare, Education, Life and Career and Music. It shows consistent performance of Llama2-13b Finetuned model for different domains.

\begin{figure*}[]
\centering
        \begin{subfigure}
                \centering
                \includegraphics[width=0.47\linewidth]{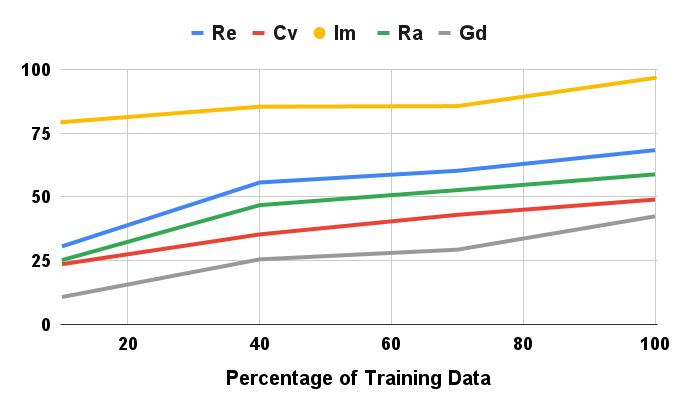}
                \label{fig:r11}
        \end{subfigure}%
                \begin{subfigure}
                \centering
                \includegraphics[width=.47\linewidth]{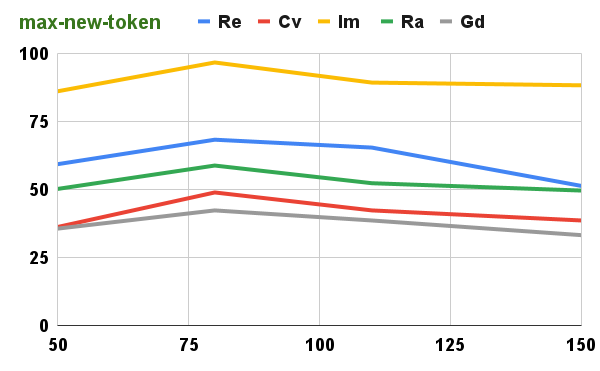}
                \label{fig:cv1}
        \end{subfigure}%
        \vspace{-3mm}
        \caption{GPT4 Criteria Performance comparison of Llama2-13b-FT model w.r.t. training data variation (left) and max-new-token size (right)} 
        \vspace{-6mm}
        \label{fig:ablation1}
\end{figure*}

\paragraph{Different Evaluation Parameter Settings}
We evaluate outcomes of various models with different parameter setting during GPT4 critique - \textit{max-new-token} and \textit{temperature}. Best results are obtained when max-new-token size is 80 (as shown in Fig. \ref{fig:ablation1}) and GPT4 critique's temperature is 0.0. 

\noindent \textbf{Varying Training Size Dataset:} 
To understand the effect of training data size on the performance, we vary the OASUM Domain-Wise Split training data for the Llama2-13b model - taking 10\%, 40\% and 70\% of the initial training data, and finetune the Llama2-13b model with same parameter and hyper-parameter settings and the five criterias of GPT4-Critique outcome (in \%) are shown in Fig \ref{fig:ablation1}. We see that with increasing the dataset size, the performance of Llama2-13b improves in terms of different GPT4 critique metrics: Relevance (Re), Coverage (Cv), Impurity (Im), Rating (Ra) and Goodness (Gd). Even at 40\% of the dataset, the model is able to achieve a decent performance. It shows the effectiveness of the Llama2-13b model. It also infers that even with very little amount of data with 10\% Llama2-13b can able to generate appropriate aspect based summary.

\noindent \textbf{Assessment of Llama2-13b Model}
To further investigate the potency of Llama2-13b-FT model, we extract 50 different OASUM articles and provide aspect based summaries of OASUM (Ground Truth vs Llama2) to two annotators (with domain knowledge and proficiency in English) to label: (i) whether Llama2-13b-FT is better, (ii) Both are Good, (iii) Ground Truth is better and (iv) Both are bad. We found that 20\% cases Llama2-13b-FT is better, 50\% cases both are good, 24\% cases ground truth is better and 6\% cases both are bad. So, overall Llama2 provides 70\% good summaries.  

These findings reinforce the claim of superiority of finetuning approach and utilization of LLM as an alternative evaluation criteria. They also show that the approach is robust to variations in type, domain, and quantity of datasets for the given task.

\section{Conclusion}
In this paper, we addressed the ever-growing challenge of efficiently extracting key insights from voluminous documents in the digital age. We explored the potential of fine-tuning large language models (LLMs) to enhance the performance of aspect-based summarization task.
Our work centered around fine-tuning open-source foundation LLMs, including Llama2, Gemma, Mistral, and Aya, on aspect-based summarization datasets. We hypothesized that this approach would enable these models to excel at identifying and extracting information relevant to user-specified aspects within a document, ultimately leading to superior quality aspect-based summaries.

Through a comprehensive evaluation framework, we compared the performance of fine-tuned LLMs against state-of-the-art aspect-based summarization methods and vanilla counterparts of the fine-tuned LLMs, and demonstrated significant improvement in quality of generated summaries as a result of fine-tuning. 
Our findings not only contribute towards the advancement of aspect-based summarization techniques but also hold significant implications for the broader field of NLP. By demonstrating the effectiveness of fine-tuning LLMs for targeted information extraction tasks like aspect-based summarization, we open doors for further exploration and potential applications in various NLP domains requiring focused information retrieval and summarization, ultimately empowering users to navigate the ever-expanding sea of information with greater efficiency and precision.

\section*{Limitations}
Our datasets are not multilingual and multimodal. We plan to capture aspects involving multimodal content, such as images or videos, limiting their comprehensiveness. LLMs may face challenges in adapting to domain-specific jargon, resulting in less informative summaries for aspects containing specialized terminology. So, we need to explore how to correct these - which we aim to do as a part of future work. 

\section*{Ethics Statement}
Our work does not reveal any personal sensitive information and we use publicly available benchmarked datasets and models in different contexts.

\bibliography{colm2024_conference}
\bibliographystyle{colm2024_conference}

\appendix

\section*{Appendix}

\section{Prompts} \label{appendix : prompts}

We use prompting in two stages - finetune-inference and critique. There are two kinds of prompts - system prompt and user prompt.

\subsection{Finetune and Inference prompt}

\textbf{system}: You are an AI assistant who is to generate the summary of a textual document specific to a certain aspect.\\
\textbf{user prompt} - Summarize the textual document given below from the perspective of {aspect}:\\ 
$\#\#\#$ Document: {document} \\

\subsection{Critique}

\textbf{system}: You are an AI assistant who is to evaluate the summary of a textual document specific to a certain aspect. You need to return a score between 0 and 1 reflecting the quality of the generated summary based on some criteria.\\
\textbf{user}:You are given a textual document and the corresponding summary of the document generated from the respective of an aspect \{aspect\} predicted by a language model as follows. \\
Document: \{document\} \\ 
Ground truth summary : \{label summary\} \\ 
Summary with respect to an aspect \{aspect\}: \{model generated summary\} \\  
Evaluate the above aspect based summary for the document in terms of each of the following criteria and return only a score between 0 and 1 without any explanation:
\begin{itemize}
    \item The extent to which the generated summary is relevant to a specific aspect \{aspect\} based summary of the document. 
    \item The extent to which the generated aspect-based summary correctly covers all the important key points described in the aspect \{aspect\} based summary of the document.  
    \item The extent to which the summary does not contain information specific to all other possible aspects \{aspect\_set\_in\_a\_domain - aspect\} based summary. 
    \item Rate the summary from the point of view of the aspect – whether the summary is good, average, or bad. A good summary effectively captures the essential points, presenting them clearly and concisely. It maintains accuracy, encourages reader engagement, and serves as a compelling introduction to the content. An average summary conveys the main points but may lack some clarity or detail, presenting a decent overview without standing out in terms of conciseness or precision. It provides a basic understanding but might benefit from a more refined focused summary fails to accurately convey the main points, containing inaccuracies or misinterpretations. It is either overly verbose or lacks coherence, making it difficult for the reader to grasp the core information effectively. 
    \item Calculated summary from the point of view of the aspect [Good/Bad/Average] [Calculated from 4 with the help of manual annotation]
\end{itemize}

\section{Time and GPU} \label{time-gpu}

We experiment on 80GB A100 GPU with GPU clock cycle 210 MHz. The finetuning and inference time of our finetuned models are in Table \ref{tab:time-gpu}.

\begin{table}[!ht]
    \centering
    \begin{tabular}{|c|c|c|}\hline
         Model & Finetune Time & Inference Time\\
         \hline
         Llama2-7b & 22 hrs & 2hrs 10 mins \\\hline
         Llama2-13b & 44 hrs & 3hrs 10 mins\\\hline
         Mistral-7b & 36 hrs & 2hrs 44 mins\\\hline
         Aya & 38 hrs &  2hrs 40mins\\\hline
         Gemma & 50 hrs &  4hrs 40mins\\\hline
    \end{tabular}
    \caption{Model Training Time [using 80GB A100 GPU]}
    \label{tab:time-gpu}
\end{table}

\section{Examples}  \label{appendix: examples}

\begin{figure*}[!ht]
    \centering
    \includegraphics{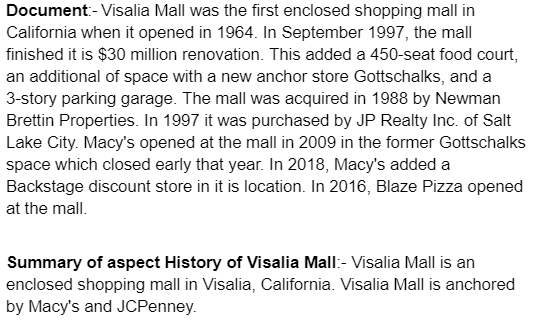}
    \caption{OASUM summary example snapshot}
    \label{fig:aspect_summ_example}
\end{figure*}

\begin{figure*}[!ht]
    \centering
    \includegraphics{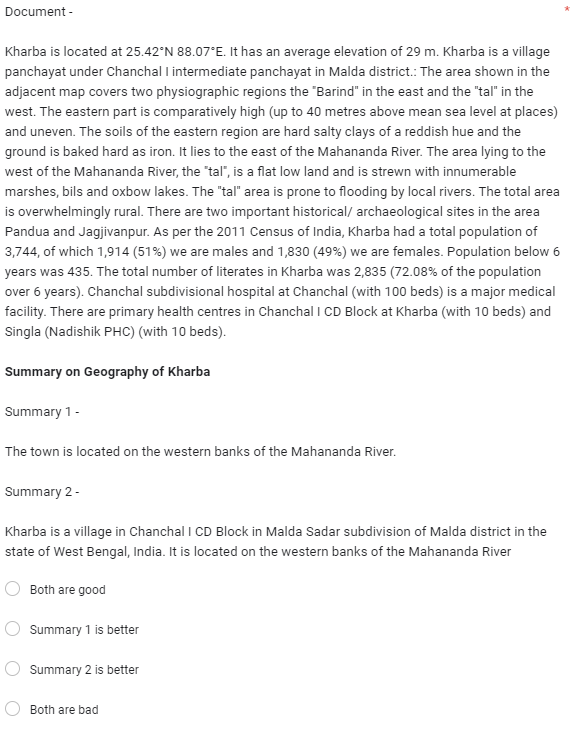}
    \caption{original summary and Llama2-13b finetune comparison experiment example snapshot}
    \label{fig:exp_1_set_1_exp}
\end{figure*}

An examples of OASUM aspect based summary is shown in Fig \ref{fig:aspect_summ_example} . Example of human annotation interface is shown in Fig \ref{fig:exp_1_set_1_exp}.

\end{document}